\documentclass[11pt,letterpaper]{article}

\usepackage[letterpaper,left=1.15in,right=1.15in,top=1.05in,bottom=1.05in]{geometry}
\usepackage[T1]{fontenc}
\usepackage[utf8]{inputenc}
\usepackage[english]{babel}

\usepackage{amsmath,amsfonts,amssymb,amsthm,mathtools}

\usepackage[mono=false]{libertine}
\usepackage[libertine]{newtxmath}
\usepackage[scaled=0.95]{inconsolata}
\usepackage{microtype}
\usepackage{graphicx}
\usepackage{booktabs}
\usepackage{multirow}
\usepackage{array}
\usepackage{tabularx}
\usepackage{wrapfig}
\usepackage{nicefrac}
\usepackage[table,dvipsnames,svgnames,x11names]{xcolor}
\usepackage{pifont}
\usepackage{xspace}
\usepackage[most]{tcolorbox}
\usepackage{enumitem}
\setlist{leftmargin=1.5em,itemsep=2pt,topsep=3pt,parsep=0pt}
\usepackage{placeins}
\usepackage{caption}
\usepackage{subcaption}
\usepackage{authblk}
\usepackage[explicit]{titlesec}
\usepackage{fancyhdr}
\usepackage{lastpage}

\PassOptionsToPackage{authoryear,round}{natbib}
\usepackage{natbib}

\usepackage[bookmarks=true,bookmarksopen=true,hidelinks]{hyperref}

\definecolor{narrared}     {HTML}{7A1022}
\definecolor{narrareddeep} {HTML}{5B0A18}
\definecolor{steelblue}    {HTML}{2E86AB}
\definecolor{mutedpurple}  {HTML}{6B4C9A}
\definecolor{warmred}      {HTML}{D1495B}
\definecolor{insightorange}{HTML}{E67E22}
\definecolor{remarkgray}   {HTML}{7F8C8D}
\definecolor{tableheadbg}  {HTML}{F4EEEF}
\definecolor{tablestripebg}{HTML}{FBF8F9}
\definecolor{tableaccentbg}{HTML}{F9EAEE}
\definecolor{tableaccentrule}{HTML}{D9B6BC}
\definecolor{tableline}    {HTML}{DDD4D7}
\definecolor{liftpositive} {HTML}{2E7D5C}
\definecolor{liftnegative} {HTML}{B23A48}

\hypersetup{
    colorlinks  = true,
    citecolor   = steelblue,
    linkcolor   = mutedpurple,
    urlcolor    = warmred,
    pdfborder   = {0 0 0}
}

\titleformat{\section}
  {\large\bfseries\scshape\color{narrareddeep}}
  {\thesection}{0.6em}{#1}[]
\titleformat{\subsection}
  {\normalsize\bfseries\color{narrareddeep}}
  {\thesubsection}{0.55em}{\itshape #1}[]
\titleformat{\subsubsection}
  {\normalsize\itshape\color{narrareddeep}}
  {\thesubsubsection}{0.5em}{#1}[]
\titlespacing*{\section}      {0pt}{2.6ex plus 0.8ex minus 0.4ex}{1.0ex}
\titlespacing*{\subsection}   {0pt}{2.0ex plus 0.7ex minus 0.3ex}{0.7ex}
\titlespacing*{\subsubsection}{0pt}{1.5ex plus 0.5ex minus 0.2ex}{0.5ex}

\renewcommand\Affilfont{\small\itshape\color{black!75}}
\makeatletter
\renewcommand\AB@affilsepx{,\enspace\protect\Affilfont}
\makeatother

\newcommand{\runningtitle}{MemoHarness}
\fancypagestyle{firststyle}{%
  \fancyhf{}%
  \fancyfoot[C]{\footnotesize\color{black!60}\thepage}%
}

\newcommand{\method}{\textsc{MemoHarness}\xspace}

\newcommand{\narraemph}[1]{\textcolor{narrareddeep}{\textbf{#1}}}
\newcommand{\methodlabel}[1]{\noindent\textbf{\textcolor{narrareddeep}{#1}}}
\newcommand{\tablehead}[1]{\textbf{\textcolor{narrareddeep}{#1}}}
\newcommand{\dimtag}[1]{{\setlength{\fboxsep}{1.6pt}\colorbox{tableheadbg}{\scriptsize\textsf{\textbf{\textcolor{narrareddeep}{#1}}}}}}
\newcommand{\dimcell}[3]{%
  \begin{tabular}[c]{@{}c@{\hspace{0.38em}}l@{}}
    \dimtag{#1} & \textbf{\textcolor{narrareddeep}{#2}}\\[-1pt]
                 & \textbf{\textcolor{narrareddeep}{#3}}
  \end{tabular}%
}
\newcommand{\dimcellone}[2]{%
  \begin{tabular}[c]{@{}c@{\hspace{0.38em}}l@{}}
    \dimtag{#1} & \textbf{\textcolor{narrareddeep}{#2}}
  \end{tabular}%
}
\newcommand{\stagecell}[2]{%
  \begin{tabular}[c]{@{}c@{}}
    \textcolor{steelblue}{#1}\\[-1pt]
    \textcolor{steelblue}{#2}
  \end{tabular}%
}

\newcommand{\defop}[2]{\textit{\textcolor{black!82}{#1}}\par\vspace{1pt}{\ttfamily\footnotesize\textcolor{warmred}{#2}}}
\newcolumntype{L}[1]{>{\raggedright\arraybackslash}p{#1}}
\newcolumntype{M}[1]{>{\raggedright\arraybackslash}m{#1}}
\newcolumntype{C}[1]{>{\centering\arraybackslash}m{#1}}

\newcommand{\logo}[1]{\raisebox{-0.20em}{\includegraphics[height=1.05em]{figures/logos/#1.png}}}

\newtheorem{definition}{Definition}[section]

\newtcolorbox{insightbox}[1][]{%
  enhanced, breakable,
  colback=white,
  colframe=white,
  borderline west={2.5pt}{0pt}{insightorange},
  sharp corners, boxrule=0pt,
  fonttitle=\bfseries\color{insightorange},
  attach title to upper={.\quad},
  title={Insight},
  #1
}

\newtcbox{\rqtag}[1][]{%
  on line,
  colback=tableheadbg,
  colframe=tableaccentrule,
  boxrule=0.4pt,
  arc=2pt,
  left=2.5pt,right=2.5pt,top=1pt,bottom=1pt,
  boxsep=0pt,
  fontupper=\bfseries\footnotesize\color{narrareddeep},
  #1
}

\title{\method: Agent Harnesses That Learn from Experience}

\author[1]{Yue Huang\textsuperscript{\textdagger}}
\author[1]{Wenjie Wang\textsuperscript{\textdagger}}
\author[1]{Han Bao\textsuperscript{\textdagger}}
\author[2]{Yuchen Ma}
\author[1]{Xiaonan Luo}
\author[3]{Yi Nian}
\author[1]{Haomin Zhuang}
\author[1]{Zheyuan Liu}
\author[3]{Yue Zhao}
\renewcommand{\thefootnote}{\fnsymbol{footnote}}

\author[1]{Xiangliang Zhang\thanks{Corresponding author: \texttt{xzhang33@nd.edu}.}}

\affil[1]{University of Notre Dame}
\affil[2]{LMU Munich}
\affil[3]{University of Southern California}

\newcommand{\equalcontributionnote}{%
  \textsuperscript{\textdagger}\,These authors contributed equally to this work.}

\newcommand{\paperstatus}{Preprint. Under review.}
\newcommand{\paperurl}{https://github.com/HowieHwong/MemoHarness}

\newcommand{\theabstract}{%
An agent harness is the external control layer that turns a base LLM into an executable agent by managing context, tools, orchestration, memory, decoding, and output handling. While harness design strongly affects agent behavior, most automatic improvement methods optimize narrower artifacts such as prompts, pipelines, or workflows, and deployed agents usually reuse a single global harness for all cases. We introduce \method, an adaptive harness optimization framework that learns from its own executions. \method decomposes the harness into six editable control dimensions, stores per-case diagnoses and distilled global patterns in a dual-layer experience bank, and adapts the learned harness to each test case using retrieved experience without test-time labels, feedback, or additional search. In our evaluation across shell-agent, code-generation, and analytical-reasoning benchmarks, \method improves over the fixed harnesses we compare against and shows selective transfer to unseen suites and base models. Its additional context can also remain cost-competitive when much of the retrieved experience is cacheable. These results provide evidence that execution experience is a practical substrate for building agent harnesses that are more adaptive than a single static configuration, while leaving broader claims about statistical robustness and component attribution to future work.%
}

\makeatletter
\newcommand{\makecoverpage}{%
  \thispagestyle{firststyle}%
  \begin{center}
    {\bfseries\fontsize{18pt}{22pt}\selectfont\color{narrareddeep}\@title\par}
    \vspace{0.95em}
    {\@author\par}
    \vspace{0.55em}
    {\footnotesize\itshape\color{black!70}\paperstatus\par}
    \vspace{0.25em}
    {\small\href{\paperurl}{\paperurl}\par}
    \ifx\equalcontributionnote\@empty\else
      \vspace{0.30em}%
      {\footnotesize\color{black!70}\equalcontributionnote\par}%
    \fi
  \end{center}

  \vspace{1.1em}

  \begin{tcolorbox}[
      enhanced, breakable,
      width=\textwidth,
      colback=tableheadbg,
      colframe=tableheadbg,
      boxrule=0pt,
      sharp corners,
      left=14pt, right=14pt, top=10pt, bottom=10pt,
  ]
    {\bfseries\color{narrareddeep}\large Abstract}\par
    \vspace{0.45em}
    {\small\theabstract\par}
  \end{tcolorbox}

  \vspace{1.2em}
}
\makeatother

\begin{document}

\makecoverpage

\renewcommand{\thefootnote}{\arabic{footnote}}
\setcounter{footnote}{0}

\section{Introduction}
\label{sec:introduction}

The performance of an LLM agent depends not only on the underlying
language model but on the entire \emph{agent harness} that surrounds
it. By \emph{agent harness}, we mean the external control layer that
turns a base LLM into an executable agent by specifying how context is
constructed, which tools and retrieval channels are exposed, how
inference is orchestrated across turns, what memory is retained, and
how outputs are validated and returned. In practice, this includes the
prompt construction policy, tool and retrieval interfaces, decoding
parameters, orchestration topology, memory management, and output
handling. Practitioner experience consistently shows that harness
design alone can swing end-to-end task success by tens of percentage
points with the same base model and the same tools
\citep{lopopolo2026harness, langchain2025context}.

Despite this, automated optimization of the \emph{harness itself}
remains rare. Most automatic improvement methods optimize a narrower
artifact: prompts or instructions
\citep{fernando2023promptbreeder}, declarative LM pipelines
\citep{khattab2023dspy}, or agent workflows
\citep{zhang2024aflow}. These methods are valuable, but they do not
jointly edit the full control layer that determines context assembly,
tool access, generation policy, orchestration, memory, and output
handling. Recent work on Meta-Harness~\citep{lee2026metaharness} moves
closer by treating harness code as the object of search, but the
result is still a training-time artifact intended to be reused at
deployment. The second missing piece is therefore \emph{test-time
adaptation}: a deployed agent should be able to specialize its harness
to the particular case in front of it without test-time labels, feedback,
or a new search run.

This setting creates three challenges. First, the harness is
high-dimensional and coupled: changing retrieval can alter the useful
prompt format, decoding budget, workflow topology, memory policy, and
validation behavior. Second, benchmark scores alone are weak supervision
for search; they reveal whether a run succeeded but not which harness
dimension caused the failure or what should be transferred to future
cases. Third, test-time adaptation must avoid leakage:
it may use only the visible input and experience accumulated during
training, yet it must handle cases that differ in domain, ambiguity,
reasoning depth, retrieval needs, and output format.

We address this gap with \method, an agent harness that learns from its
own past executions. \method makes three design choices.
\textbf{(1)~Six-dimensional harness space.} We decompose the harness
along the temporal flow of inference into six editable control
surfaces (context, tool, generation, orchestration, memory, output),
turning harness search into structured editing over separable
dimensions rather than over a single opaque prompt
(\S\ref{sec:harness_space}).
\textbf{(2)~Dual-layer experience bank.} During search, \method
accumulates both per-case execution entries and distilled global
patterns about what works, what fails, and how dimensions interact, so
that each iteration becomes diagnostic rather than only score-driven
(\S\ref{sec:experience_bank}).
\textbf{(3)~Test-time case adaptation.} At test time, the
search-derived global harness $W^*$ is adapted to a case-specific
harness $W(x_j)$ by retrieving similar past cases and relevant patterns
from the bank, without any test-time feedback, gradient updates, or
extra search rounds (\S\ref{sec:test_time}). A correctness-first ranking
rule, with execution cost only as a tiebreaker, keeps the search from
drifting onto cheap-but-wrong configurations.
Figure~\ref{fig:pipeline} summarizes the full pipeline.

We evaluate \method across benchmarks spanning shell tooling, code
generation, and analytical reasoning. On the primary shell-agent
benchmark under our evaluation protocol, \method improves over the
strongest fixed-harness baseline we run ($0.806$ vs.\ $0.722$) while
reducing per-task dollar cost relative to the strongest commercial
baselines in that comparison. The learned harness also shows selective,
not uniform, transfer without retraining: it improves some unseen
evaluation suites and raises average task success across the held-out
base models we evaluate by $+0.098$. These results point to a practical
role for adaptive harnesses: execution experience can be reused to
specialize the control layer case by case, although larger-scale studies
and finer component ablations are needed to establish the robustness and
source of the gains.

\methodlabel{Contributions.}
The main contributions of this work are as follows:
\begin{itemize}[leftmargin=1.35em, itemsep=0pt, topsep=1pt, parsep=0pt, partopsep=0pt]
  \item A \emph{structured harness optimization framework} that
  decomposes a monolithic agent harness into six editable control
  surfaces and pairs them with a dual-layer experience bank, so that
  search accumulates reusable diagnostic knowledge rather than only
  scalar scores (\S\ref{sec:harness_space} and~\S\ref{sec:experience_bank}).
  \item A \emph{test-time adaptation mechanism} that adapts
  the search-derived global harness to each new case using retrieved
  successes, failures, and global patterns, without test-time feedback
  (\S\ref{sec:test_time}).
  \item Empirical results across in-domain, cross-dataset, and
  cross-model evaluations showing improved task success under our
  protocol, selective positive transfer across unseen evaluation suites,
  gains on the evaluated base-model families, and favorable cost behavior
  when retrieved context is reusable through caching
  (\S\ref{sec:experiments}).
\end{itemize}

\section{Method}
\label{sec:method}

\begin{figure}[t]
\centering
\includegraphics[width=\columnwidth]{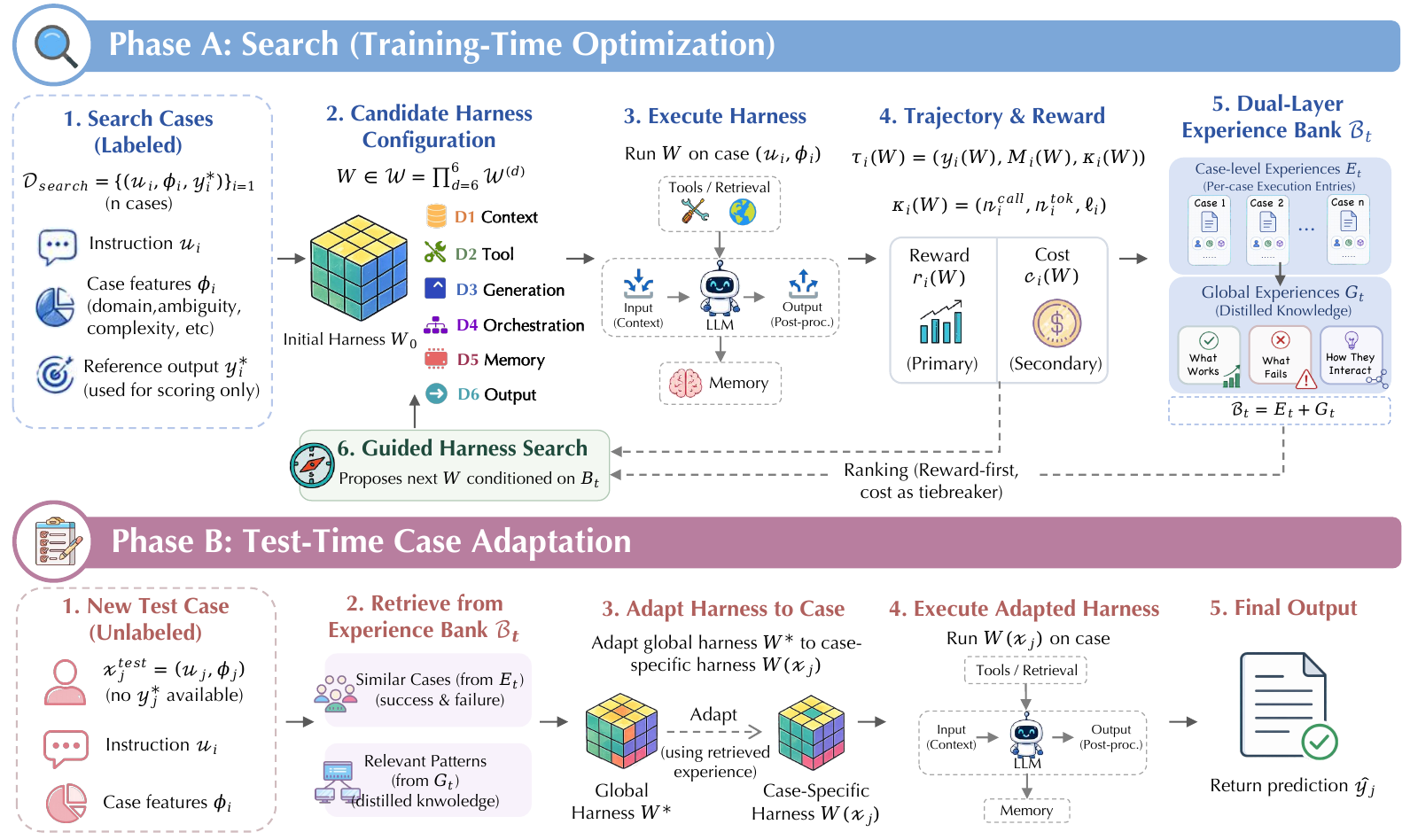}
\vspace{-15pt}
\caption{Overview of \method. \textbf{Phase~A (Search, training-time
optimization)} runs a guided search over the six-dimensional harness
space $\mathcal{W} = \prod_{d=1}^{6}\mathcal{W}^{(d)}$: each candidate
harness $W$ is executed on labeled search cases, scored by a
correctness-first reward $r_i(W)$ with token usage $c_i(W)=n_i^{\mathrm{tok}}$ as tiebreaker, and
the resulting trajectories are stored in a dual-layer experience bank
$\mathcal{B}_t = (E_t, G_t)$ of per-case entries and distilled global
patterns; the search policy $\pi$ then proposes the next $W$
conditioned on $\mathcal{B}_t$. \textbf{Phase~B (Test-Time Case
Adaptation)} takes an unlabeled test case $x_j^{\text{test}}$, retrieves
similar cases and relevant patterns from $\mathcal{B}_t$, and maps the
global harness $W^*$ to a case-specific harness $W(x_j)$, which is then
executed to produce the final prediction $\hat{y}_j$.}
\label{fig:pipeline}
\vspace{-15pt}
\end{figure}

\methodlabel{Overview.}
We study \emph{adaptive harness optimization} for a fixed base language model.
A \emph{harness} is the surrounding system that determines what information is
retrieved, how it is assembled, how model calls are orchestrated, and how raw
outputs are post-processed.
Rather than treating optimization as only a training-time search for one
reusable global artifact, \method learns a \emph{global base harness} during
search and applies a \emph{case-specific adaptation rule} at test time.
The method combines a structured six-dimensional harness space with a dual-layer
experience bank and follows a \narraemph{correctness-first} principle: the
primary reward determines ranking, while lower execution cost serves only as a
secondary tiebreaker. Figure~\ref{fig:pipeline} summarizes the full pipeline,
which decomposes into a training-time search phase and a test-time
case-adaptation phase.

\subsection{Motivation}
\label{sec:motivation}

LLM harnesses are usually deployed as a single benchmark-level configuration:
one prompt template, decoding policy, tool interface, and orchestration pattern
is selected and then reused for all cases. This is brittle when cases differ in
domain, required reasoning depth, retrieval needs, or output format. A harness
that is strong on average can still be systematically mismatched to particular
case types.

\method addresses this mismatch by making harness search diagnostic and
case-adaptive. It decomposes the harness into six editable control dimensions,
records which dimensions explain past successes and failures, and reuses that
experience to adapt the learned global harness once for each test case.

\subsection{Problem Formulation}
\label{sec:formulation}

\begin{definition}[Benchmark Cases]
\label{def:case}
We distinguish between labeled search cases and unlabeled evaluation cases.
The search set is
\begin{equation}
\label{eq:search_set}
\mathcal{D}_{\mathrm{search}}
=
\bigl\{x_i^{\mathrm{s}}=(u_i,\; \phi_i,\; y_i^\star)\bigr\}_{i=1}^{n},
\end{equation}
where $u_i$ is the input instruction, $\phi_i$ denotes case features such as
domain, ambiguity, complexity, and whether external knowledge is required, and
$y_i^\star$ is the reference output used only for search-time scoring. The
evaluation set is
\begin{equation}
\label{eq:test_set}
\mathcal{D}_{\mathrm{test}}
=
\bigl\{x_j^{\mathrm{test}}=(u_j,\; \phi_j)\bigr\}_{j=1}^{m}.
\end{equation}
At test time, the adaptation rule may condition on $(u_j,\phi_j)$ and the
experience bank, but not on the unknown reference output.
\end{definition}

\begin{definition}[Harness Configuration]
\label{def:harness}
For each dimension $d \in \{1,\ldots,6\}$, let $\mathcal{W}^{(d)}$ denote the
set of admissible choices for that functional control surface. A harness
configuration is an element of the product space
\begin{equation}
\label{eq:harness}
W \in \mathcal{W}
=
\mathcal{W}^{(1)}
\times
\mathcal{W}^{(2)}
\times
\cdots
\times
\mathcal{W}^{(6)},
\qquad
W = \bigl(W^{(1)},\ldots,W^{(6)}\bigr),
\end{equation}
where each component $W^{(d)}$ controls one functional dimension of the
inference pipeline (see Section~\ref{sec:harness_space}).
\end{definition}

\noindent
Executing the visible part $(u_i,\phi_i)$ of a case under configuration $W$
produces a prediction $y_i(W)$. We denote the corresponding execution
trajectory by $\tau_i(W)$; it records the prediction, intermediate
model-tool trace, and lightweight runtime diagnostics:
\begin{equation}
\label{eq:trajectory}
\tau_i(W)
=
\bigl(
  y_i(W),\;
  \mathcal{M}_i(W),\;
  \kappa_i(W)
\bigr),
\qquad
\kappa_i(W)
=
\bigl(n_i^{\mathrm{call}}(W),\;
n_i^{\mathrm{tok}}(W),\;
\ell_i(W)\bigr),
\end{equation}
where $\mathcal{M}_i(W)$ includes invoked tools and intermediate outputs,
$n_i^{\mathrm{call}}(W)$ is the number of model calls,
$n_i^{\mathrm{tok}}(W)$ is total token usage, and $\ell_i(W)$ is latency.
We use total token usage as the search-time secondary cost,
\begin{equation}
\label{eq:cost}
c_i(W) = n_i^{\mathrm{tok}}(W),
\end{equation}
with lower values preferred. The remaining components of $\kappa_i(W)$,
call count and latency, are logged for diagnostic and reporting
purposes but do not enter the harness-selection rule. Monetary cost in
dollars is computed offline from token counts at the public list prices
of the base model (RQ5) and is therefore
separate from search-time selection. Given a user-specified primary
metric $R$, the search-time task reward is
\begin{equation}
\label{eq:reward}
r_i(W) = R \!\bigl(y_i(W),\; y_i^\star\bigr),
\end{equation}
with larger values preferred.

\noindent
The search problem is to identify a strong global harness $W^\star$ from
candidate executions on $\mathcal{D}_{\mathrm{search}}$. The evaluation-time
problem is to construct, for each unlabeled case $(u_j,\phi_j)$, a specialized
harness $W(x_j)$ using only test-visible information and retrieved experience.

\subsection{Six-Dimensional Harness Space}
\label{sec:harness_space}

\begin{insightbox}[title={Design rationale}]
Rather than treating the harness as a monolithic prompt or agent template, we
decompose it according to the \emph{temporal information flow} of inference.
Each dimension corresponds to a distinct functional stage, enabling targeted
diagnosis and repair while making cross-dimensional interactions explicit.
\end{insightbox}

\begin{table}[t]
\centering
\caption{Six dimensions of the \method harness space.}
\label{tab:dimensions}
\small
\setlength{\tabcolsep}{5pt}
\renewcommand{\arraystretch}{1.24}
\arrayrulecolor{tableline}
\begin{tabular}{@{}M{0.20\textwidth}C{0.20\textwidth}M{0.54\textwidth}@{}}
\toprule
\rowcolor{tableheadbg}
\tablehead{Dimension} & \tablehead{Stage} & \tablehead{Definition and example operation} \\
\midrule
\rowcolor{tablestripebg}
\dimcell{D1}{Context}{assembly}
  & \stagecell{Pre-call input}{construction}
  & \defop{Builds the model input from instructions, constraints, retrieved material, and examples.}
    {structure prompt; add demos; compress context} \\
\addlinespace[3pt]
\dimcell{D2}{Tool}{interaction}
  & \stagecell{External tool and}{retrieval use}
  & \defop{Controls when and how the harness calls external tools or retrievers.}
    {enable retrieval; set top-k; rerank evidence} \\
\addlinespace[3pt]
\rowcolor{tablestripebg}
\dimcell{D3}{Generation}{control}
  & \stagecell{Decoding}{configuration}
  & \defop{Sets the sampling and budget parameters for model generation.}
    {raise max tokens; lower temperature; sample candidates} \\
\addlinespace[3pt]
\dimcellone{D4}{Orchestration}
  & \stagecell{Workflow}{topology}
  & \defop{Chooses the sequence of model calls and intermediate reasoning steps.}
    {single call -> plan/execute/refine} \\
\addlinespace[3pt]
\rowcolor{tablestripebg}
\dimcell{D5}{Memory}{management}
  & \stagecell{Cross-call state}{persistence}
  & \defop{Determines what state persists across calls and what stale context is removed.}
    {keep state; summarize trace; drop stale context} \\
\addlinespace[3pt]
\dimcell{D6}{Output}{processing}
  & \stagecell{Post-call output}{handling}
  & \defop{Transforms raw model output into the final answer returned by the harness.}
    {extract answer; validate schema; choose fallback} \\
\bottomrule
\end{tabular}
\arrayrulecolor{black}
\vspace{-15pt}
\end{table}

\noindent
This decomposition (Table~\ref{tab:dimensions}) turns harness search into
structured editing over \emph{separable control surfaces} instead of treating
the entire harness as a single opaque object.

\subsection{Dual-Layer Experience Bank}
\label{sec:experience_bank}

\method maintains an experience bank as a typed pair
\begin{equation}
\label{eq:bank}
\mathcal{B}_t
=
\bigl(\mathcal{E}_t,\; \mathcal{G}_t\bigr),
\end{equation}
where $\mathcal{E}_t$ is the set of per-case execution entries accumulated up
to search iteration $t$, and $\mathcal{G}_t$ is the set of distilled global
patterns. We use a pair rather than a set union because entries and patterns
have different schemas.

\methodlabel{Per-case experience.}
For each search case and iteration, we store an execution entry
\begin{equation}
\label{eq:per_case}
\xi_i^{(t)} =
\Bigl(
i,\,
t,\,
\phi_i,\,
W_t,\,
\Delta_i^{(t)},\,
\tau_i(W_t),\,
r_i(W_t),\,
c_i(W_t),\,
z_i^{(t)}
\Bigr),
\end{equation}
where
\begin{equation}
\label{eq:delta}
\Delta_i^{(t)}
=
\Delta\!\bigl(W_t,\; W_i^{<t}\bigr)
\end{equation}
is the configuration delta from $W_i^{<t}$, the most recent harness previously
applied to case $i$ (or $W_0$ if no such execution exists). The diagnosis is
produced by a diagnostic operator
\begin{equation}
\label{eq:diagnostic_operator}
z_i^{(t)}
=
g\!\bigl(x_i^{\mathrm{s}},\; W_t,\; \tau_i(W_t),\; r_i(W_t)\bigr),
\end{equation}
whose output schema is
\begin{equation}
\label{eq:diagnosis}
z_i^{(t)} =
\bigl(
s_i^{(t)},\;
d_{i,\text{prim}}^{(t)},\;
\mathcal{D}_{i,\text{sec}}^{(t)},\;
a_i^{(t)}
\bigr),
\end{equation}
where $s_i^{(t)} \in \{0,1\}$ indicates success,
$d_{i,\text{prim}}^{(t)} \in \{1,\ldots,6\}\cup\{\emptyset\}$ is the primary
failure dimension, with $\emptyset$ used when no failure dimension is assigned,
$\mathcal{D}_{i,\text{sec}}^{(t)}$ are secondary contributing dimensions, and
$a_i^{(t)}$ is a natural-language analysis. We also maintain lightweight
per-case statistics, including consecutive failures, recent average reward,
reward trend, and dimension-wise failure counts.

\methodlabel{Global patterns and retrieval.}
Every $N$ search iterations, a distillation operator extracts persistent
cross-case regularities from failure clusters,
\begin{equation}
\label{eq:distill}
\mathcal{G}_t
\leftarrow
\mathcal{G}_{t-1}
\cup
\mathrm{Distill}\!\bigl(\mathcal{E}_{\leq t}\bigr),
\end{equation}
where each global pattern summarizes a recurring phenomenon, its supporting
evidence, and the expected effect of a targeted harness change. The controller
does not read the entire bank directly; instead, it issues a structured query
$q \in \mathcal{Q}$ over case features, failure statistics, iteration ranges,
and dimension-specific diagnoses, yielding a retrieved slice
\begin{equation}
\label{eq:retrieve}
\mathcal{S}_t(q)
=
\mathrm{Retrieve}\!\bigl(\mathcal{B}_t,\; q\bigr).
\end{equation}
A retrieved slice may contain global patterns, filtered per-case entries, and
aggregated statistics, which keeps the controller context bounded even as the
bank grows over time.

\subsection{Training-Time Harness Optimization}
\label{sec:training}

Training proceeds over the labeled search set
$\mathcal{D}_{\mathrm{search}}$ for $T$ iterations. We begin from a
\emph{minimal harness} $W_0$: no demonstrations,
no structured instruction scaffolding, no external tools, deterministic
single-call decoding, no cross-call memory, and raw output passthrough.

\begin{insightbox}[title={Starting from scratch}]
Initializing from a minimal harness ensures that every component of the final
configuration is \emph{justified by empirical evidence} accumulated during
search, rather than inherited from potentially suboptimal defaults.
\end{insightbox}

\noindent
At search iteration $t$, the controller forms a query $q_t$ from the current
harness and the accumulated bank, retrieves a bounded evidence slice
$\mathcal{S}_{t-1}(q_t)$, and proposes the next configuration:
\begin{equation}
\label{eq:controller}
q_t = Q\!\bigl(W_{t-1},\; \mathcal{B}_{t-1}\bigr),
\qquad
\mathcal{S}_{t-1}(q_t)
=
\mathrm{Retrieve}\!\bigl(\mathcal{B}_{t-1},\; q_t\bigr),
\qquad
W_t
=
\Pi_{\mathrm{train}}\!\bigl(W_{t-1},\; \mathcal{S}_{t-1}(q_t)\bigr).
\end{equation}
Here $Q$ is a query constructor and $\Pi_{\mathrm{train}}$ is the
retrieval-conditioned controller. The proposed harness is then executed on
every search case $x_i^{\mathrm{s}} \in \mathcal{D}_{\mathrm{search}}$ to
collect trajectories $\tau_i(W_t)$, compute rewards $r_i(W_t)$ and costs
$c_i(W_t)$, and produce diagnosis signals $z_i^{(t)}$. All resulting entries
are written to $\mathcal{E}_t$, and every $N$ rounds the system distills
persistent failure clusters into new entries in $\mathcal{G}_t$.

\methodlabel{Correctness-first selection.}
For each candidate $W_t$, we compute mean task reward and mean cost:
\begin{equation}
\label{eq:mean_rewards}
\bar{r}_t
=
\frac{1}{n}
\sum_{i=1}^{n} r_i(W_t),
\qquad
\bar{c}_t
=
\frac{1}{n}
\sum_{i=1}^{n} c_i(W_t).
\end{equation}
Let $\mathcal{C}_T=\{W_0,\ldots,W_T\}$ denote the explored candidates. If a
cost budget or resource constraint is imposed, let
$\mathcal{C}_{\mathrm{feas}}\subseteq\mathcal{C}_T$ denote the feasible
candidates; otherwise $\mathcal{C}_{\mathrm{feas}}=\mathcal{C}_T$. The final
global harness $W^\star$ is selected by \emph{lexicographic ordering}:
\begin{equation}
\label{eq:selection}
W^\star
\in
\operatorname*{\arg\max}_{\mathrm{lex},\, W_t \in \mathcal{C}_{\mathrm{feas}}}
\bigl(\bar{r}_t,\; -\bar{c}_t\bigr).
\end{equation}
\noindent
This makes the optimization objective explicit: primary reward is optimized
first, while lower token usage ($\bar{c}_t$) serves only as a secondary
tiebreaker among candidates with equal primary reward.

\subsection{Test-Time Case Adaptation Without Feedback}
\label{sec:test_time}

After training, \method enters the evaluation phase on a test set
$\mathcal{D}_{\mathrm{test}}$.
Unlike training, test-time adaptation operates \emph{without any iterative
feedback loop}.
For each unseen case $x=(u,\phi)$, the controller adapts the learned
global harness $W^\star$ using only
test-visible information and the frozen experience bank $\mathcal{B}_T$.

\methodlabel{Test-time case adaptation.}
For the current case, the controller retrieves global patterns,
feature-matched historical slices, and the most similar successful and failed
training examples. Let $\psi(u)$ denote an instruction representation, and let
$u_\xi$ denote the instruction associated with historical entry $\xi$. We score
the similarity between a test case $x=(u,\phi)$ and an entry $\xi$ by
\begin{equation}
\label{eq:similarity}
\rho_\psi(x,\xi)
=
\cos\!\bigl(\psi(u),\; \psi(u_\xi)\bigr).
\end{equation}
Let
$\mathcal{E}_T^{+}=\{\xi\in\mathcal{E}_T: s(\xi)=1\}$ and
$\mathcal{E}_T^{-}=\{\xi\in\mathcal{E}_T: s(\xi)=0\}$ denote successful and
failed historical entries. The retrieved neighborhoods are
\begin{equation}
\label{eq:neighbors}
\mathcal{N}^{+}_K(x)
=
\operatorname*{TopK}_{\xi \in \mathcal{E}_T^{+}}
\bigl[\rho_\psi(x,\xi)\bigr],
\qquad
\mathcal{N}^{-}_K(x)
=
\operatorname*{TopK}_{\xi \in \mathcal{E}_T^{-}}
\bigl[\rho_\psi(x,\xi)\bigr],
\end{equation}
where $\operatorname*{TopK}$ returns the entries with the largest scores.
Together with a feature-conditioned retrieval slice, these neighborhoods form
the test-time evidence
\begin{equation}
\label{eq:test_evidence}
\mathcal{S}_{\mathrm{test}}(x)
=
\Bigl(
\mathcal{N}^{+}_K(x),\;
\mathcal{N}^{-}_K(x),\;
\mathrm{Retrieve}\!\bigl(\mathcal{B}_T,\; Q_{\mathrm{test}}(x)\bigr),\;
\mathcal{G}_T
\Bigr).
\end{equation}
Conditioned on this evidence, the controller emits a case-specific harness
\begin{equation}
\label{eq:test_adapt}
W(x)
=
\Pi_{\mathrm{test}}\!\bigl(W^\star,\; x,\; \mathcal{S}_{\mathrm{test}}(x)\bigr),
\end{equation}
which is then executed for the current evaluation case. If evaluation labels or metrics later
become available, the resulting trajectory may optionally be appended to the
experience bank for future reuse, but no learning, re-selection, or
distillation is triggered during the current evaluation.

\methodlabel{Relation to global harness search.}
The central distinction of \method is the separation of
\emph{global optimization} from \emph{instance-level adaptation}.
Training identifies a robust global harness under \narraemph{correctness-first}
selection, while evaluation uses the experience bank to specialize that harness
to the requirements of each individual case.
This allows simple cases to remain lightweight, while retrieval-heavy,
multi-step, or format-sensitive cases can invoke richer orchestration only when
the retrieved evidence warrants it.

\section{Experiments}
\label{sec:experiments}

We organize the evaluation around five research questions. RQ1 asks
whether adaptive harness optimization improves absolute task success
against existing agent harnesses. RQ2 examines whether search produces
stable improvements over iterations and across task families. RQ3
tests whether learned harness changes transfer to unseen evaluation
suites. RQ4 asks whether the same harness transfers across base models.
RQ5 measures whether the added retrieval and adaptation machinery is
cost-effective at inference time.

\subsection{Setup}
\label{sec:setup}

We evaluate \method on three task families: long-horizon shell agency
(\texttt{Terminal-Bench}), single-shot code generation
(\texttt{LiveCodeBench}), and multi-step financial reasoning
(\texttt{FinanceAgent}). On \texttt{Terminal-Bench}, we compare against
four released or benchmark-provided harness baselines under a shared
evaluation protocol; when a baseline is not a model-agnostic harness, we
use the closest reproducible configuration available in that system
rather than claiming a pure scaffold-only transplant. For transfer, we
also evaluate the learned harness on six external suites and six
additional base models. We
report mean task success rate averaged over repeated runs, using the
validation-selected harness rather than the in-training peak.
Appendix~\ref{app:experimental_details} gives benchmark, baseline,
model, and cost-accounting details.

\subsection{Results}
\label{sec:results}

\noindent\rqtag{RQ1}\hspace{0.45em}\textbf{Does adaptive harness optimization improve task success?}

\begin{wrapfigure}{r}{0.44\columnwidth}
\vspace{-15pt}
\centering
\includegraphics[width=0.44\columnwidth]{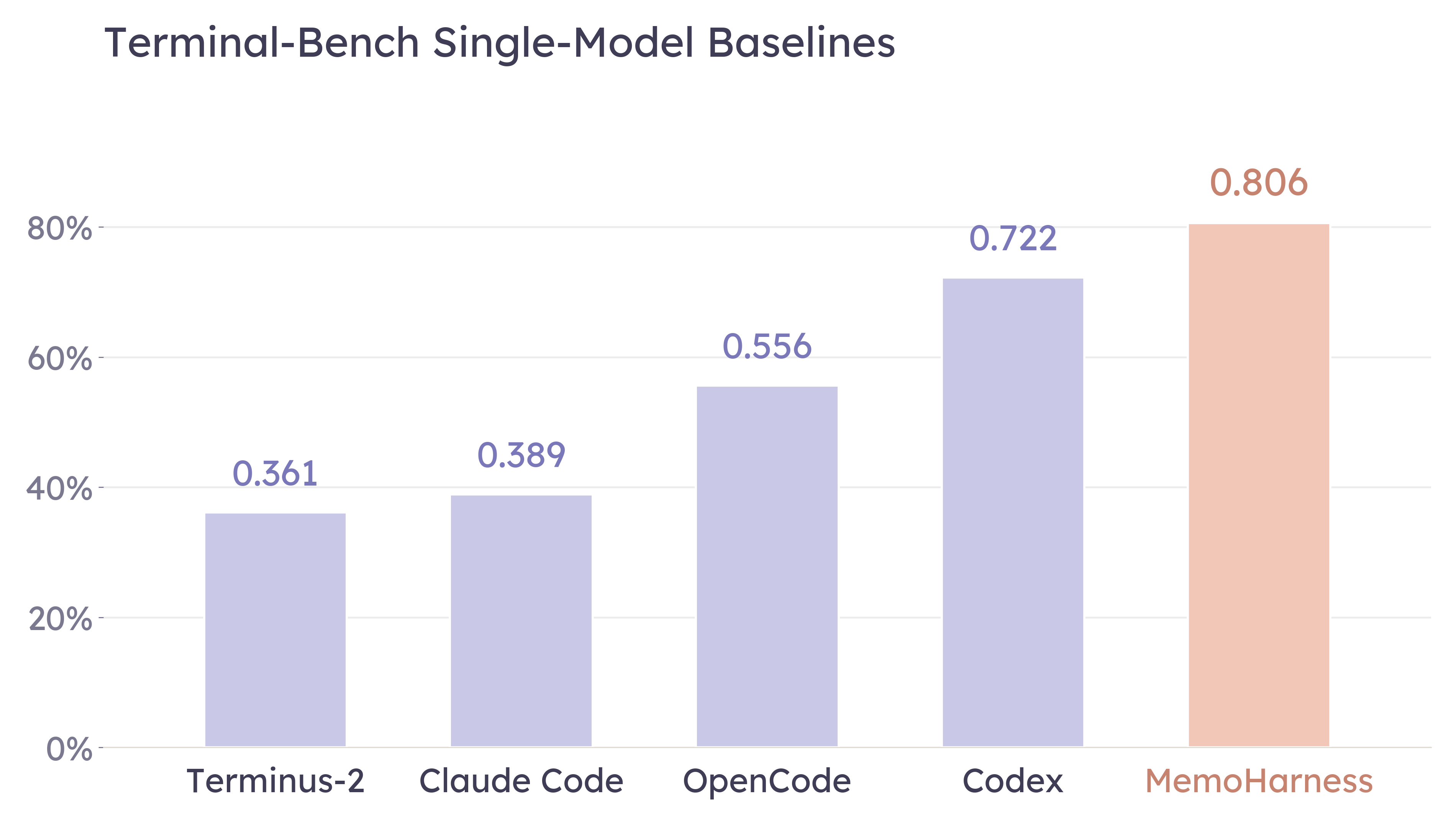}
\vspace{-15pt}
\caption{Baseline comparison on \texttt{Terminal-Bench}. Higher is
better; \method (rightmost, using \texttt{GPT-5.3-Codex}) reaches the
highest mean of $0.806$.}
\label{fig:tb_baselines}
\vspace{-15pt}
\end{wrapfigure}

Figure~\ref{fig:tb_baselines} compares \method to four released or
benchmark-provided harness baselines on \texttt{Terminal-Bench}.
\method reaches $0.806$, improving over the strongest baseline,
\texttt{Codex}, by $+0.084$, and over the other
baselines by $+0.250$ to $+0.445$. For baselines whose underlying
generator can be held fixed, these comparisons are closer to isolating
the surrounding harness. For product-specific baselines, they should be
read as comparisons to the closest released system configuration under
our protocol rather than as pure model-controlled scaffold ablations.

The comparison is intentionally stringent in the sense that
\texttt{Codex} is already a specialized terminal-oriented harness, not
a weak prompt-only baseline. The remaining headroom is consistent with
the view that end-to-end agent performance is meaningfully shaped by
harness control decisions such as context assembly, workflow topology,
memory, and output handling. In this setting, \method appears to recover
part of that headroom by using search-time execution experience instead
of relying only on manually authored defaults.

\noindent\rqtag{RQ2}\hspace{0.45em}\textbf{How does harness quality evolve during search?}

\begin{figure}[t]
\centering
\includegraphics[width=\columnwidth]{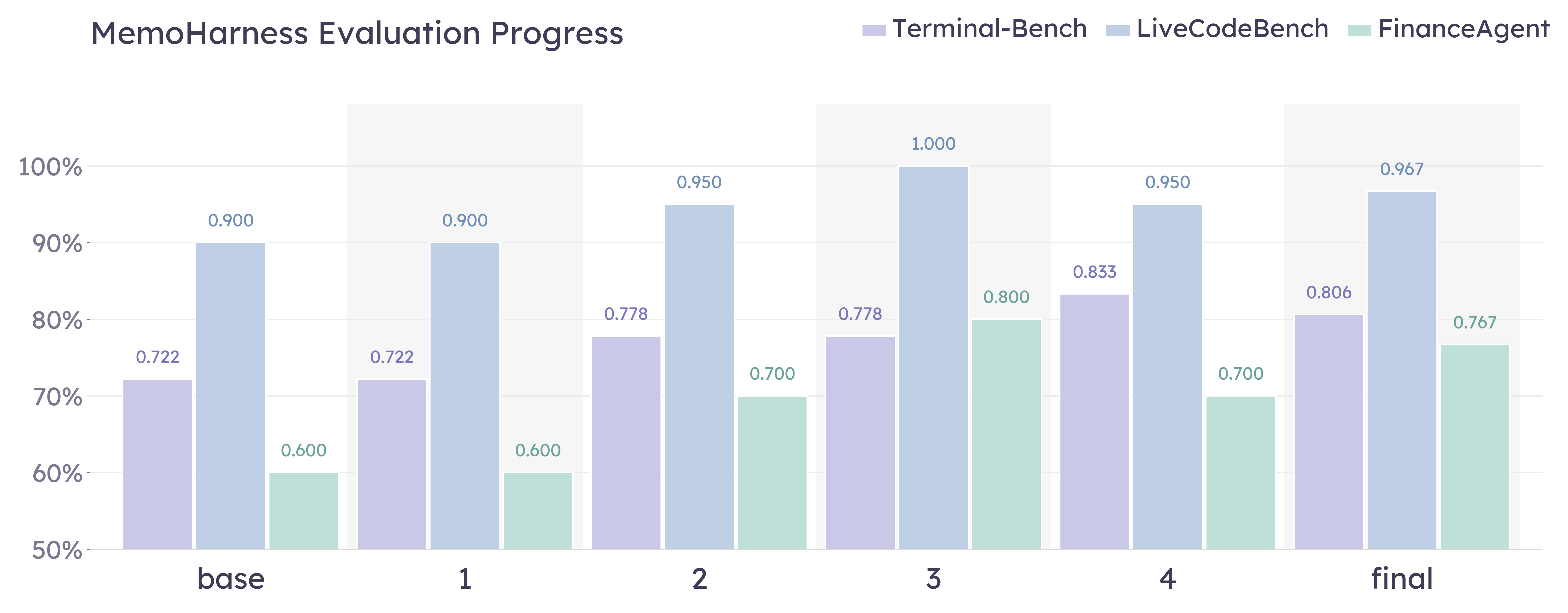}
\vspace{-15pt}
\caption{Harness quality at six checkpoints (\texttt{base}, four
intermediate iterations, and the final selected harness) across three
benchmarks. The final score is higher than \texttt{base} on every
benchmark, but it is not always equal to the in-training peak:
\texttt{LiveCodeBench} touches $1.000$ at iteration~3, and
\texttt{Terminal-Bench} reaches $0.833$ at iteration~4 because the
shipped harness is selected by validation rather than by peeking at the
held-out evaluation split.}
\label{fig:progress}
\vspace{-11.8pt}
\end{figure}

\begin{figure}[t]
\centering
\includegraphics[width=\columnwidth]{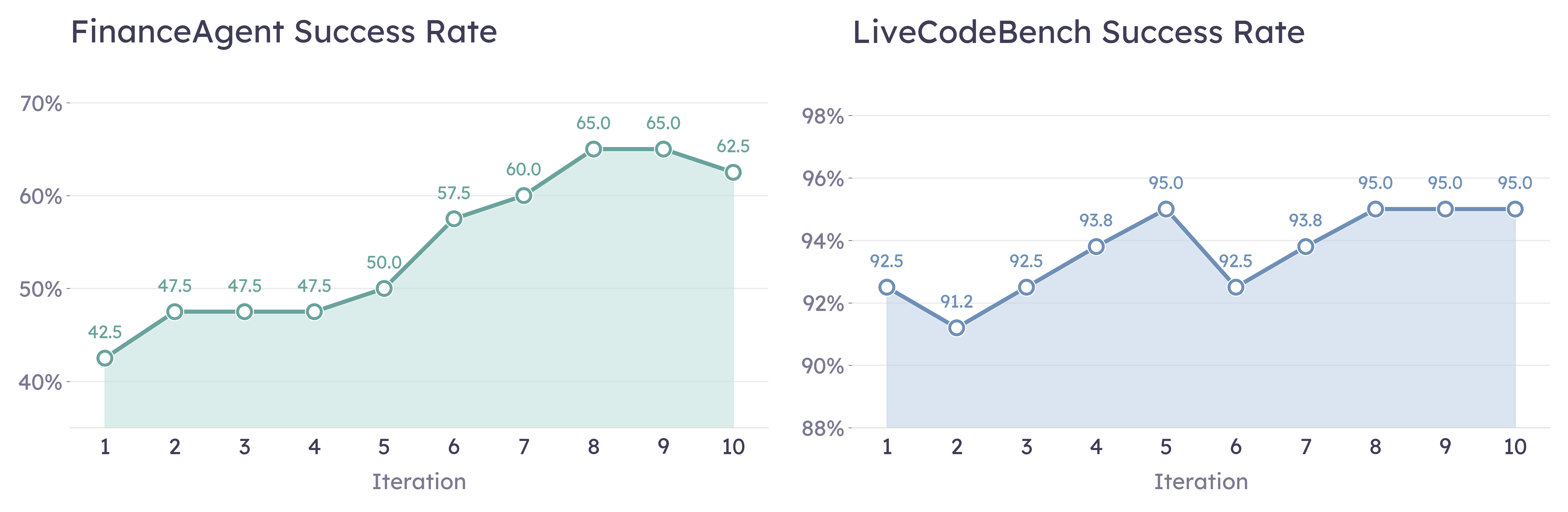}
\vspace{-15pt}
\caption{Per-iteration success rate on \texttt{FinanceAgent} (left) and
\texttt{LiveCodeBench} (right) over $10$ search rounds.
\texttt{FinanceAgent} continues to benefit from additional iterations
(rising from $42.5\%$ to a $65.0\%$ peak around iterations $8$ and~$9$),
whereas \texttt{LiveCodeBench} saturates almost immediately near the
base-model ceiling and oscillates within a $\sim$$4$pt band.}
\label{fig:iteration}
\vspace{-15pt}
\end{figure}

To understand how the harness improves over the course of optimization,
we examine its quality at intermediate checkpoints across all three
benchmarks (Figure~\ref{fig:progress}) and the per-iteration trajectory
on the two non-\texttt{Terminal-Bench} benchmarks
(Figure~\ref{fig:iteration}).

Figure~\ref{fig:progress} shows that the final selected harness
improves over \texttt{base} on all three benchmarks:
$0.722\!\to\!0.806$ on \texttt{Terminal-Bench}, $0.900\!\to\!0.967$
on \texttt{LiveCodeBench}, and $0.600\!\to\!0.767$ on
\texttt{FinanceAgent}. The largest gain appears on
\texttt{FinanceAgent}, where the base harness leaves the most room;
\texttt{LiveCodeBench} is near saturation and improves less. The final
checkpoint is sometimes below the in-training peak because selection is
validation-based rather than peak-test-score based.

Figure~\ref{fig:iteration} shows that \texttt{FinanceAgent}
continues to improve across search rounds, reaching $65.0\%$ near iterations $8$
and~$9$, while \texttt{LiveCodeBench} remains in a narrow
$91.2\%$--$95.0\%$ band. Thus harness search helps most on longer-horizon
agentic workloads, while still giving smaller gains in near-saturated
single-shot code generation.

The different trajectories are useful diagnostically. On
\texttt{FinanceAgent}, later iterations still discover beneficial edits,
suggesting that multi-step analytical tasks expose diverse failure modes
that can be repaired over time. On \texttt{LiveCodeBench}, by contrast,
the base model already solves most cases, so search mostly operates in a
low-headroom regime where small validation differences matter. This
supports the use of validation-based selection and motivates the
operation-level breakdown in Appendix~\ref{app:op_lift}, which asks
which atomic edits are associated with reward improvements.

\noindent\rqtag{RQ3}\hspace{0.45em}\textbf{Do learned harness changes transfer to unseen suites?}

A learned harness is more convincing if it transfers not only across
search checkpoints and base models, but also across evaluation suites
that were never used during search. To test this, we take the final
harness learned on each source benchmark under
\texttt{GPT-5.3-Codex} and evaluate it on the six external
suites described in Section~\ref{sec:setup}. For each source
benchmark, we compare against the same untuned \texttt{Codex} base
harness; all reported metrics are the suites' native higher-is-better
scores.

\begin{table*}[t]
\centering
\caption{Cross-dataset generalization of search-derived harnesses under
the shared base model \texttt{GPT-5.3-Codex}. The first row is the
shared \texttt{Codex} baseline; each \texttt{MemoHarness} row is
learned from the listed search source.
\texttt{HEFix}, \texttt{RG-Easy}, and \texttt{SWE-Pro} abbreviate
\texttt{HumanEvalFix}, \texttt{Reasoning-Gym-Easy}, and
\texttt{SWE-Bench Pro}. Higher is better in every column.}
\label{tab:cross_dataset_results}
\footnotesize
\setlength{\tabcolsep}{3pt}
\renewcommand{\arraystretch}{1.14}
\arrayrulecolor{tableline}
\begin{tabular*}{\textwidth}{@{\extracolsep{\fill}}llcccccc@{}}
\toprule
\rowcolor{tableheadbg}
\tablehead{Search source} & \tablehead{Harness}
& \tablehead{\texttt{MMMLU}}
& \tablehead{\texttt{HEFix}}
& \tablehead{\texttt{StrongR}}
& \tablehead{\texttt{RG-Easy}}
& \tablehead{\texttt{Law}}
& \tablehead{\texttt{SWE-Pro}} \\
\midrule
\rowcolor{tablestripebg}
\textit{Shared baseline}
& \texttt{Codex} & 0.818 & 1.000 & 0.879 & 0.947 & 0.675 & 0.706 \\
\arrayrulecolor{tableaccentrule}\cmidrule(lr){1-8}\arrayrulecolor{tableline}
\rowcolor{tableaccentbg}
\texttt{Terminal-Bench}
& \textbf{\textcolor{narrareddeep}{\texttt{MemoHarness}}}
& \textbf{\textcolor{narrareddeep}{0.848}} & 1.000
& \textbf{\textcolor{narrareddeep}{0.909}} & 0.947 & \textbf{\textcolor{narrareddeep}{0.676}}
& \textbf{\textcolor{narrareddeep}{0.765}} \\
\rowcolor{tableaccentbg}
\texttt{FinanceAgent}
& \textbf{\textcolor{narrareddeep}{\texttt{MemoHarness}}} & 0.818 & 1.000
& \textbf{\textcolor{narrareddeep}{0.909}} & 0.947 & \textbf{\textcolor{narrareddeep}{0.682}} & 0.706 \\
\rowcolor{tableaccentbg}
\texttt{LiveCodeBench}
& \textbf{\textcolor{narrareddeep}{\texttt{MemoHarness}}}
& \textbf{\textcolor{narrareddeep}{0.879}} & 1.000
& \textbf{\textcolor{narrareddeep}{0.909}} & 0.947 & 0.669 & 0.706 \\
\bottomrule
\end{tabular*}
\arrayrulecolor{black}
\vspace{-11pt}
\end{table*}

Table~\ref{tab:cross_dataset_results} shows positive but selective
transfer. The \texttt{Terminal-Bench} harness gives the broadest lift,
improving \texttt{MMMLU} by $+0.030$, \texttt{StrongReject} by
$+0.030$, and \texttt{SWE-Bench Pro} by $+0.059$. The
\texttt{LiveCodeBench} harness improves \texttt{MMMLU} and
\texttt{StrongReject}, while the \texttt{FinanceAgent} harness mainly
improves \texttt{StrongReject} and \texttt{LawBench}. Saturated suites
such as \texttt{HumanEvalFix} and \texttt{Reasoning-Gym-Easy} show no
movement.

These results argue against two extreme interpretations. The learned
harness is not a universally dominant prompt template: some source
benchmarks produce conservative or domain-specific edits, and
\texttt{LawBench} remains mixed. At the same time, the observed gains
are not confined to the source benchmark. The strongest transfer appears
when the source search task is long-horizon and tool-centric, suggesting
that some learned control decisions, such as more robust instruction
following or software-task structure, may survive a change in evaluation
suite. This interpretation remains provisional because we do not
separately ablate every component of the transferred harness.

\noindent\rqtag{RQ4}\hspace{0.45em}\textbf{Does the learned harness transfer across base models?}

A learned harness is most useful if it transfers across base models
without retraining. We take the harness produced by searching with
\texttt{GPT-5.3-Codex} and apply it directly to six additional models
spanning four families.

\begin{wraptable}{r}{0.55\columnwidth}
\vspace{-15pt}
\centering
\caption{Cross-model transfer on \texttt{Terminal-Bench}. Higher is better.}
\label{tab:transfer_results}
\footnotesize
\setlength{\tabcolsep}{4pt}
\renewcommand{\arraystretch}{1.18}
\arrayrulecolor{tableline}
\begin{tabular}{@{}L{0.50\linewidth}C{0.16\linewidth}C{0.26\linewidth}@{}}
\toprule
\rowcolor{tableheadbg}
\tablehead{Model} & \tablehead{Base} & \tablehead{\texttt{MemoHarness}} \\
\midrule
\rowcolor{tablestripebg}
\logo{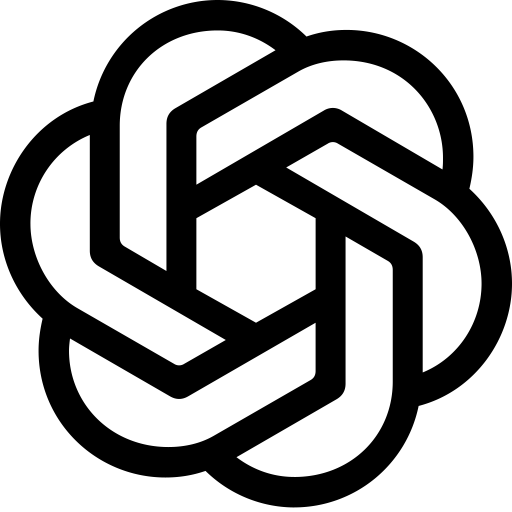}~\texttt{GPT-5.3-Codex} & 0.722 & \textbf{\textcolor{narrareddeep}{0.806}} \\
\logo{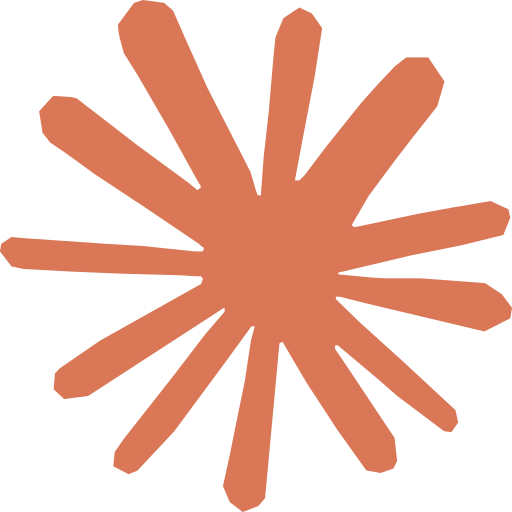}~\texttt{Claude-Sonnet-4.6} & 0.530 & \textbf{\textcolor{narrareddeep}{0.583}} \\
\rowcolor{tablestripebg}
\logo{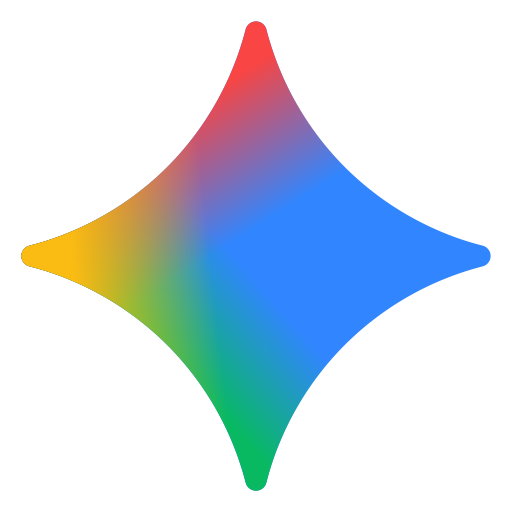}~\texttt{Gemini-3.1-Pro} & 0.611 & \textbf{\textcolor{narrareddeep}{0.694}} \\
\logo{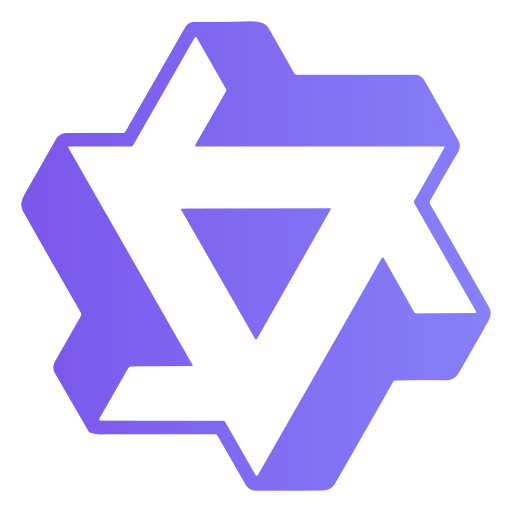}~\texttt{Qwen3.5-397B-A17B} & 0.444 & \textbf{\textcolor{narrareddeep}{0.528}} \\
\rowcolor{tablestripebg}
\logo{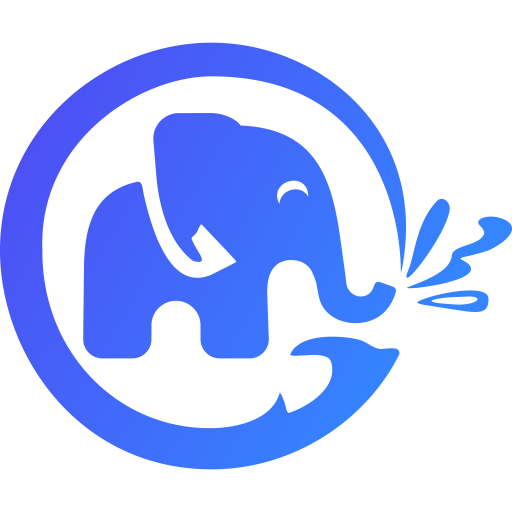}~\texttt{GLM-5} & 0.500 & \textbf{\textcolor{narrareddeep}{0.733}} \\
\logo{openai}~\texttt{GPT-4.1} & 0.500 & \textbf{\textcolor{narrareddeep}{0.538}} \\
\rowcolor{tablestripebg}
\logo{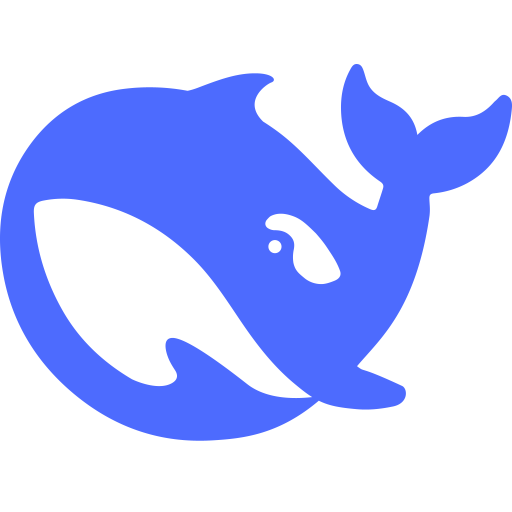}~\texttt{DeepSeek-V3.2} & 0.333 & \textbf{\textcolor{narrareddeep}{0.444}} \\
\bottomrule
\end{tabular}
\arrayrulecolor{black}
\vspace{-15pt}
\end{wraptable}

Table~\ref{tab:transfer_results} shows that every evaluated model
improves over its base setting in our transfer study. The mean gain is
$+0.098$, ranging from $+0.038$ on \texttt{GPT-4.1} to $+0.233$ on
\texttt{GLM-5}. The source model, \texttt{GPT-5.3-Codex}, retains the
same $+0.084$ in-domain improvement reported in
Figure~\ref{fig:tb_baselines}.

The transfer pattern is important because the harness was not
re-searched for each model. Improvements on both proprietary and
open-weight models suggest, but do not by themselves prove, that the
learned changes are not purely model-specific prompt quirks. They are
consistent with a more portable execution policy for how an LLM should
gather context, invoke tools, manage intermediate state, and finalize
outputs on terminal tasks. The smaller gain on \texttt{GPT-4.1}
suggests that stronger or already well-calibrated base settings may
leave less room for harness intervention, but this pattern should be
confirmed on larger model and task sets.

\noindent\rqtag{RQ5}\hspace{0.45em}\textbf{Is adaptive harnessing cost-effective?}

Table~\ref{tab:cost_results} reports token usage and dollar cost on the
\texttt{Terminal-Bench} 18-task held-out evaluation split under
\texttt{GPT-5.3-Codex}.
Dollar costs are computed offline from public list prices and are
reported for comparison only; harness selection uses raw token usage as
the secondary tiebreaker (Eq.~\ref{eq:selection}).

\begin{table}[t]
\centering
\caption{Cost analysis on \texttt{Terminal-Bench} (18-task held-out
evaluation split) with
\texttt{GPT-5.3-Codex} as the shared base model. Token counts are reported in
millions; lower dollar cost is better.}
\vspace{-6pt}
\label{tab:cost_results}
\footnotesize
\setlength{\tabcolsep}{4pt}
\renewcommand{\arraystretch}{1.14}
\arrayrulecolor{tableline}
\begin{tabular}{@{}L{0.22\textwidth}rrrrr@{}}
\toprule
\rowcolor{tableheadbg}
\tablehead{Framework} & \tablehead{Input (M)} &
\tablehead{Cached (M)} & \tablehead{Non-cached (M)} & \tablehead{Output (M)} &
\tablehead{Cost (\$)} \\
\midrule
\rowcolor{tablestripebg}
\logo{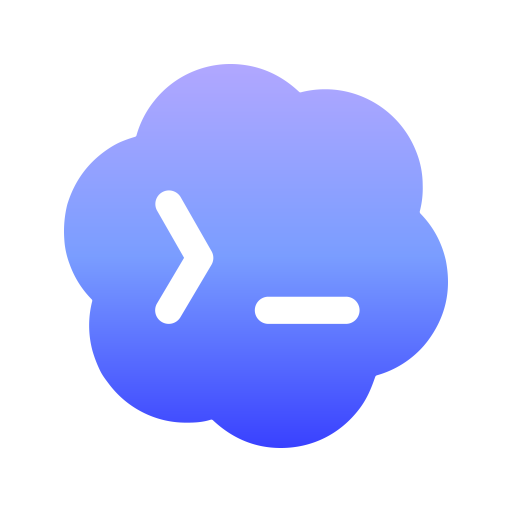}~\texttt{Codex} & 8.23 & 4.33 & 3.90 & 0.19 & 10.28 \\
\texttt{Terminus} & 3.96 & 0.94 & 3.03 & 0.09 & 6.68 \\
\rowcolor{tablestripebg}
\logo{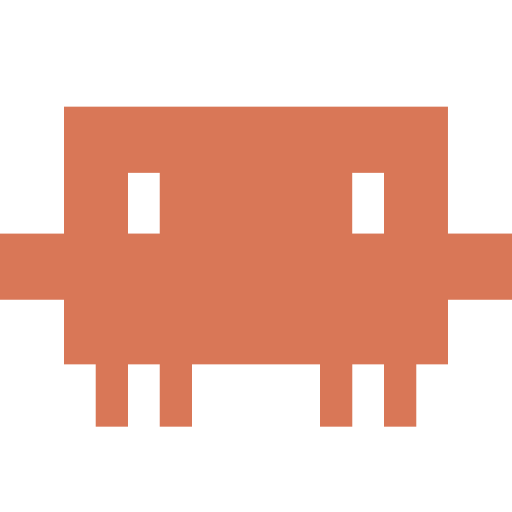}~\texttt{Claude Code} & 7.32 & 3.11 & 4.21 & 0.11 & 9.51 \\
\logo{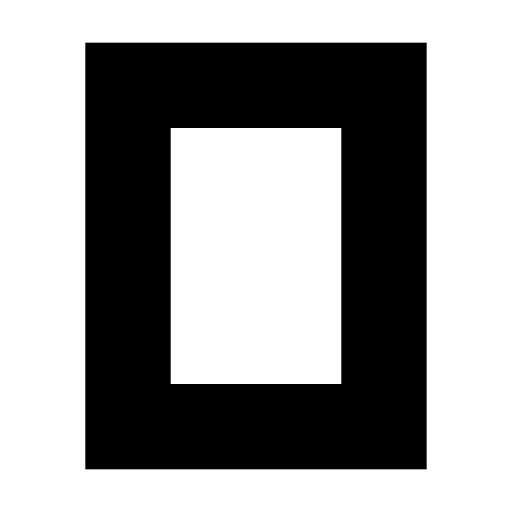}~\texttt{OpenCode} & 5.48 & 5.07 & 0.41 & 0.05 & 2.34 \\
\arrayrulecolor{tableaccentrule}\cmidrule(lr){1-6}\arrayrulecolor{tableline}
\rowcolor{tableaccentbg}
\textbf{\textcolor{narrareddeep}{\texttt{MemoHarness}}}
& \textbf{\textcolor{narrareddeep}{14.18}}
& \textbf{\textcolor{narrareddeep}{13.32}}
& \textbf{\textcolor{narrareddeep}{0.86}}
& \textbf{\textcolor{narrareddeep}{0.22}}
& \textbf{\textcolor{narrareddeep}{6.89}} \\
\bottomrule
\end{tabular}
\arrayrulecolor{black}
\vspace{-25pt}
\end{table}

\method uses more raw input tokens because it retrieves from the
experience bank, but most of that context is cached in our evaluation
($13.32$M of $14.18$M input tokens). Its total reported cost is
therefore $\$6.89$, below \texttt{Codex} and \texttt{Claude Code} while
achieving higher task success under this cost-accounting protocol.
\texttt{Terminus} and \texttt{OpenCode} remain cheaper, but at
substantially lower accuracy. The cost comparison should be interpreted
with this caching assumption in mind.

\vspace{-10pt}

\section{Conclusion}
\label{sec:conclusion}

\vspace{-5pt}

We presented \method, an agent harness that turns past executions into
reusable experience for future cases. By combining a structured
six-dimensional harness space, a dual-layer experience bank, and
test-time case adaptation, \method makes harness search both diagnostic
and reusable. The results suggest that improving the control layer
around an LLM can be a practical complement to model scaling and manual
harness engineering. Future work should study fully unsupervised search,
larger-scale validation, more detailed component attribution, and online
experience accumulation across deployments.

\bibliographystyle{plainnat}
\bibliography{references}

\appendix

\section*{Appendix Contents}
\label{app:contents}
\begin{tcolorbox}[
  enhanced,
  colback=tableheadbg,
  colframe=tableaccentrule,
  boxrule=0.5pt,
  arc=2pt,
  left=7pt,
  right=7pt,
  top=6pt,
  bottom=6pt
]
\small
\renewcommand{\arraystretch}{1.18}
\begin{tabular*}{\linewidth}{@{\extracolsep{\fill}}C{0.08\linewidth}L{0.70\linewidth}r@{}}
\textbf{\textcolor{narrareddeep}{\ref{sec:limitations}}}
  & \hyperref[sec:limitations]{Limitations}
  & \textcolor{remarkgray}{p.~\pageref{sec:limitations}} \\
\textbf{\textcolor{narrareddeep}{\ref{app:implementation_details}}}
  & \hyperref[app:implementation_details]{Implementation details}
  & \textcolor{remarkgray}{p.~\pageref{app:implementation_details}} \\
\textbf{\textcolor{narrareddeep}{\ref{app:experimental_details}}}
  & \hyperref[app:experimental_details]{Experimental details}
  & \textcolor{remarkgray}{p.~\pageref{app:experimental_details}} \\
\textbf{\textcolor{narrareddeep}{\ref{app:models}}}
  & \hyperref[app:models]{Models used}
  & \textcolor{remarkgray}{p.~\pageref{app:models}} \\
\textbf{\textcolor{narrareddeep}{\ref{app:comparison_scope}}}
  & \hyperref[app:comparison_scope]{Comparison scope}
  & \textcolor{remarkgray}{p.~\pageref{app:comparison_scope}} \\
\textbf{\textcolor{narrareddeep}{\ref{sec:related}}}
  & \hyperref[sec:related]{Related work}
  & \textcolor{remarkgray}{p.~\pageref{sec:related}} \\
\textbf{\textcolor{narrareddeep}{\ref{app:op_lift}}}
  & \hyperref[app:op_lift]{Operation-level diagnostic}
  & \textcolor{remarkgray}{p.~\pageref{app:op_lift}}
\end{tabular*}
\end{tcolorbox}
\vspace{-6pt}

\section{Limitations}
\label{sec:limitations}

This study has several limitations. First, the primary
\texttt{Terminal-Bench} evaluation uses an 18-task held-out split, and
the present version reports point estimates rather than confidence
intervals or significance tests. The results should therefore be read as
evidence under a fixed evaluation protocol, not as a full statistical
characterization of harness performance. Second, not every baseline is a
pure scaffold transplant with the same underlying model and runtime
surface. Where systems expose different model, tool, or product-level
interfaces, our comparisons are necessarily system-level comparisons to
the closest reproducible released configuration. Third, the current
experiments do not fully isolate every component of \method: in
particular, we do not separately ablate the experience bank, global
patterns, and case-specific test-time adaptation in all settings. Fourth,
the cost analysis depends on the observed cacheability of retrieved
experience in our runs; deployments with lower cache reuse may see a
different cost profile. Finally, the reference implementation
instantiates the controller and diagnostic operators with practical
heuristics. This makes the system reproducible and inspectable, but
future work should study learned or otherwise more general controllers,
larger held-out splits, and online accumulation of experience across
deployments.

\section{Implementation Details}
\label{app:implementation_details}

This appendix describes how the abstract components of
Section~\ref{sec:method}, namely the harness space, the diagnostic
operator, and the cost proxy, are realized in our reference
implementation. None of these choices alters the algorithmic structure
of \method; they specify how the abstract objects are concretely
instantiated in the experiments reported in
Section~\ref{sec:experiments}.

\paragraph{Harness representation.}
A harness configuration is defined in the main text as a tuple
$W = (W^{(1)}, \ldots, W^{(6)})$ over the six functional dimensions
(Section~\ref{sec:harness_space}). In the implementation, $W$ is
materialized as a \emph{harness bundle}: a structured policy file that
records the current D1 to D6 configuration, paired with lightweight
textual scaffolding specifying the agent's operating rules, a
persistent playbook, and the distilled memory currently in scope. An
edit to a harness therefore corresponds to editing both the structured
dimension summary and the accompanying instructions that instantiate
those choices at execution time, keeping the textual scaffolding in
sync with the typed dimension state.

\paragraph{Initialization.}
All experiments reported in this paper initialize from the minimal
harness $W_0$ in which demonstrations, retrieval, structured
scaffolding, cross-call memory, and output validators are all disabled.
The implementation also supports resuming from a previously archived
bundle when one is present, but this deployment-time convenience is not
used for the reported benchmark results. In both modes, the search
procedure optimizes over the same six-dimensional space
$\mathcal{W} = \prod_{d=1}^{6} \mathcal{W}^{(d)}$.

\paragraph{Diagnostic instantiation.}
The diagnostic operator $g$ from Section~\ref{sec:experience_bank} is
instantiated with verifier-grounded signals from the execution
environment. Rather than requiring a full causal explanation for each
failure, the implementation maps observable evidence, including verifier
outcomes, exception types, timeouts, missing artifacts, command
failures, and execution traces, to one of the six dimensions
$d \in \{1, \ldots, 6\}$, yielding a per-case diagnosis $z_i^{(t)}$
that is intentionally coarse and serves as a stable optimization
signal. More persistent structure is recovered at the global layer:
when repeated failures accumulate, they are distilled into cross-case
entries in $\mathcal{G}_t$ together with supporting evidence and a
dimension-level repair suggestion.

\paragraph{Cost proxy.}
The main text instantiates the search-time cost as total token usage,
$c_i(W) = n_i^{\mathrm{tok}}(W)$, leaving the remaining components of
$\kappa_i(W)$, call count and latency, for diagnostic reporting
only (Eq.~\ref{eq:cost}). The implementation follows this exactly:
token usage is the only cost quantity that enters the lexicographic
selection rule (Eq.~\ref{eq:selection}), and dollar cost is computed
offline at report time from token counts and the public list prices
of the base model. Token usage was chosen because it is available
consistently across runs and correlates directly with inference cost;
call count and latency are logged for debugging and for the
breakdowns in RQ5, but never affect
selection. This preserves the correctness-first ordering described in
the main text while avoiding dependence on less stable runtime
measurements.

\paragraph{Global pattern distillation.}
The experience bank $\mathcal{B}_t = (\mathcal{E}_t, \mathcal{G}_t)$
pairs per-case execution entries with distilled global patterns
(Section~\ref{sec:experience_bank}). In the implementation,
distillation runs both on a fixed schedule (every $N$ search rounds,
as in the main text) and opportunistically when repeated failures
accumulate on the same case. The distiller consumes recent failure
histories, reward trends, dimension assignments, and compact
configuration snapshots, and emits a bounded number of new global
patterns into $\mathcal{G}_t$. Capping the number of newly emitted
patterns per round keeps the controller's prompt budget bounded
across iterations while preserving recurring lessons from prior
executions.

\section{Experimental Details}
\label{app:experimental_details}

\paragraph{Benchmarks.}
We use three benchmarks chosen to span distinct task families.
\texttt{Terminal-Bench} is a long-horizon shell-agent benchmark
involving multi-step tool use, file editing, and process management. We
use an 80/20 (8:2) train/evaluation split of the full 89-task suite,
yielding an 18-task held-out evaluation split for absolute performance
and cost reporting. \texttt{LiveCodeBench} consists of recent
competitive-programming problems and stresses single-shot code
generation rather than multi-step agency. \texttt{FinanceAgent}
requires multi-step analytical reasoning over financial documents and
tool calls, exercising the longer-horizon analytical end of the
workload spectrum. Together the three benchmarks cover shell tooling,
code generation, and analytical reasoning, three regimes in which
harness design plays qualitatively different roles.

For cross-dataset evaluation, we additionally test the learned
harnesses on six external suites: \texttt{MMMLU}
\citep{hendryckstest2021}, \texttt{HumanEvalFix} from OctoPack
\citep{muennighoff2024octopack}, \texttt{StrongReject}
\citep{souly2024a}, \texttt{Reasoning-Gym-Easy}
\citep{stojanovski2026reasoning}, \texttt{LawBench}
\citep{fei-etal-2024-lawbench}, and \texttt{SWE-Bench Pro}
\citep{deng2026swebench}. These suites cover knowledge-intensive
question answering, code repair, safety refusal behavior, verifiable
reasoning, legal reasoning, and long-horizon software engineering.

\paragraph{Baselines.}
On \texttt{Terminal-Bench}~\citep{merrill2026terminalbench} we compare
against four agent-harness frameworks: OpenAI's
\texttt{Codex}~\citep{openai2025codexcli}, Anthropic's \texttt{Claude
Code}~\citep{anthropic2025claudecode}, the open-source
\texttt{OpenCode}~\citep{sst2024opencode}, and \texttt{Terminus}, the
neutral test-bed agent that ships with the benchmark. We run each
baseline using its released default configuration or the closest
reproducible configuration available under the shared evaluation
protocol. Where the framework permits a model-agnostic generator swap,
we use \texttt{GPT-5.3-Codex}; where it does not, we treat the result as
a comparison to the released system rather than as a pure scaffold-only
ablation. Of the four,
\texttt{Codex} is the strongest baseline in our setup and serves as
the most demanding reference point for \method.

\paragraph{Models and protocol.}
The harness search uses \texttt{GPT-5.3-Codex} as the base model. For
cross-model transfer, we evaluate the search-derived harness, without
any further training or retraining, on six other base models from four
families: \texttt{Claude-Sonnet-4.6}, \texttt{Gemini-3.1-Pro},
\texttt{Qwen3.5-397B-A17B}, \texttt{GLM-5}, \texttt{GPT-4.1}, and
\texttt{DeepSeek-V3.2}. We report mean task success rate per benchmark,
averaged over repeated runs. Because reported values are run averages,
they are not constrained to integer multiples of the 18 held-out
\texttt{Terminal-Bench} tasks. The harness used at evaluation time is
the one selected by validation at the end of search, not the
in-training peak shown in Figures~\ref{fig:progress}
and~\ref{fig:iteration}. For cost reporting, token counts are averaged
over repeated runs on the \texttt{Terminal-Bench} evaluation split;
dollar costs use the public list prices of \texttt{GPT-5.3-Codex} for
cached input, non-cached input, and output tokens.

\paragraph{Hyperparameters and splits.}
The Controller--Bank loop runs for $T=10$ outer iterations. Global
pattern distillation uses a dual trigger: it runs after $M=5$ new bank
entries or after $N=3$ consecutive failures on the same case, whichever
occurs first. At each round, the controller receives a compact bank
summary containing $K_{\mathrm{succ}}=10$ recent successes and
$K_{\mathrm{fail}}=10$ recent failures. We do not use semantic retrieval
for these summaries ($D2.\texttt{top\_k}=0$). Dataset splits use a fixed
80/20 (8:2) random partition with seed $42$, so repeated runs and the
held-out evaluation pass use identical task IDs. Generation is
deterministic in all reported runs: temperature $0.0$, top-$p=1.0$,
candidate count $1$, and maximum generation budget $8192$ tokens.

\section{Models Used}
\label{app:models}

Table~\ref{tab:models} lists the seven base models evaluated in this
paper. The harness search itself uses \texttt{GPT-5.3-Codex} as the
source model (Section~\ref{sec:setup}); the remaining six models
participate only in the cross-model transfer study
in RQ4, where they execute the
search-derived harness with no further training. ``Access''
distinguishes proprietary models served only behind an API from
models whose weights have been publicly released.

\begin{table}[t]
\centering
\caption{Base models used in this paper. ``Access'' indicates whether
the model's weights are publicly released or served only via a
proprietary API. \texttt{GPT-5.3-Codex} is the source model used
during harness search; the remaining six are evaluated only at test
time, as described in RQ4.}
\label{tab:models}
\footnotesize
\setlength{\tabcolsep}{8pt}
\renewcommand{\arraystretch}{1.20}
\arrayrulecolor{tableline}
\begin{tabular}{@{}lll@{}}
\toprule
\rowcolor{tableheadbg}
\tablehead{Model} & \tablehead{Provider} & \tablehead{Access} \\
\midrule
\rowcolor{tablestripebg}
\logo{openai}~\texttt{GPT-5.3-Codex}              & OpenAI           & Proprietary (API) \\
\logo{claude-color}~\texttt{Claude-Sonnet-4.6}    & Anthropic        & Proprietary (API) \\
\rowcolor{tablestripebg}
\logo{gemini-color}~\texttt{Gemini-3.1-Pro}       & Google DeepMind  & Proprietary (API) \\
\logo{openai}~\texttt{GPT-4.1}                    & OpenAI           & Proprietary (API) \\
\rowcolor{tablestripebg}
\logo{qwen-color}~\texttt{Qwen3.5-397B-A17B}      & Alibaba          & Open weights \\
\logo{chatglm-color}~\texttt{GLM-5}               & Zhipu AI         & Open weights \\
\rowcolor{tablestripebg}
\logo{deepseek-color}~\texttt{DeepSeek-V3.2}      & DeepSeek         & Open weights \\
\bottomrule
\end{tabular}
\arrayrulecolor{black}
\vspace{-15pt}
\end{table}

\section{Comparison Scope}
\label{app:comparison_scope}

Meta-Harness~\citep{lee2026metaharness} is the closest prior system to
our training-time search setting because it also treats the harness as
the object of optimization. We discuss it as related work rather than as
a direct empirical baseline because a public implementation suitable for
our evaluation stack was not available to us at the time our experiments
were finalized (early 2026); reference code released later (e.g., the
Stanford IRIS Lab repository) was not incorporated into this version of
the paper. Reimplementing the method from the paper would introduce
substantial design choices about proposer interfaces, candidate
execution, trace storage, and benchmark adapters, making the resulting
comparison difficult to attribute cleanly. We therefore restrict direct
empirical comparisons to runnable harnesses with released implementations
or evaluation substrates, and compare Meta-Harness qualitatively in
Section~\ref{sec:related}.

\section{Related Work}
\label{sec:related}

\subsection{Optimization for Agents}
\label{sec:related_opt}

Recent work studies how to automatically improve prompts, inference policies, and
agentic workflows rather than treating them as fixed hand-written artifacts.
Early agentic prompting and tool-use methods such as ReAct
\citep{yao2023react} and Toolformer \citep{schick2023toolformer} established
that performance improves substantially when inference is augmented with
actions, tools, and external feedback. Planning-oriented inference methods such
as Tree of Thoughts \citep{yao2023treeofthoughts} and Language Agent Tree Search
\citep{zhou2023lats} further demonstrated that search over intermediate reasoning
states can improve deliberative decision making.

Prompt and instruction optimization subsequently became an explicit objective.
OPRO treats the language model itself as an optimizer over instruction space
\citep{yang2023opro}; ProTeGi performs prompt optimization using textual
gradients and beam search \citep{pryzant2023automatic}; and Promptbreeder
evolves both task prompts and mutation prompts \citep{fernando2023promptbreeder}.
A complementary line of work improves generation behavior through iterative
self-feedback: Self-Refine repeatedly generates feedback and revisions
\citep{madaan2023selfrefine}, while Reflexion stores verbal reflections in
memory to improve subsequent trials \citep{shinn2023reflexion}.
However, these methods optimize \emph{within a single prompt turn} and do not
consider the broader system-level configuration that surrounds the model.

System-level optimization methods move beyond a single prompt. DSPy formulates
language-model pipelines as declarative programs that can be compiled against a
task metric \citep{khattab2023dspy}; MIPRO extends this line by optimizing
instructions and demonstrations for multi-stage language-model programs
\citep{opsahlong2024mipro}; and TextGrad casts optimization of compound AI
systems as textual backpropagation \citep{yuksekgonul2024textgrad}. For agent
workflows, AutoFlow automatically generates natural-language workflow programs
\citep{li2024autoflow}, and AFlow optimizes code-represented agent workflows
with search and execution feedback \citep{zhang2024aflow}. Closest to our
setting, Meta-Harness searches directly over harness code
\citep{lee2026metaharness}, and AlphaEvolve demonstrates the power of
evolutionary code-level improvement under evaluator feedback
\citep{novikov2025alphaevolve}.
\emph{Across this line of work, optimization is primarily a pre-deployment
procedure over prompts, programs, workflows, or harness code. Less explored is
harness-level adaptation to the specific requirements of each individual test
case at deployment time}. \method addresses this gap through its
dual-layer experience bank and test-time case adaptation.

\subsection{Harness Engineering}
\label{sec:related_harness}

Recent practitioner work increasingly argues that agent performance depends as
much on the surrounding harness as on the underlying model. Anthropic's
\emph{Building Effective Agents} emphasizes simple compositional workflows,
careful context management, and deliberate tool design in production systems
\citep{anthropic2024effectiveagents}. Their subsequent engineering note on tool
design argues that tool naming, descriptions, response shaping, and evaluation
directly affect both task success and efficiency \citep{anthropic2025tools}.
LangChain's writing on \emph{context engineering} similarly frames reliability
as a problem of deciding what should be written, selected, compressed, and
isolated across long agent trajectories \citep{langchain2025context}. OpenAI's
\emph{Harness engineering} post extends this view to software agents, stressing
repository legibility, structured in-repo knowledge, and feedback loops as
first-class engineering artifacts rather than incidental implementation details
\citep{lopopolo2026harness}.

Academic work has only recently begun to formalize these intuitions.
Meta-Harness makes the harness itself the object of search
\citep{lee2026metaharness}, while Natural-Language Agent Harnesses externalize
harness behavior as editable natural-language artifacts executed by a shared
runtime \citep{pan2026naturallanguageharnesses}. \texttt{Terminal-Bench}
provides a realistic command-line evaluation substrate on which different
software-agent harnesses can be compared \citep{merrill2026terminalbench}. In
parallel, SWE-agent and OpenHands demonstrate that interface design,
environment access, and execution scaffolding are central determinants of agent
performance rather than incidental implementation details
\citep{yang2024sweagent,wang2024openhands}.
\emph{While this emerging literature establishes harness engineering as a
first-class concern, we are not aware of prior work that accumulates reusable
diagnostic experience across search iterations and leverages it for case-level
adaptation at test time}. This is the contribution that \method
makes.

\section{Operation-Level Diagnostic on Adjacent-Iteration Transitions}
\label{app:op_lift}

A scalar score curve only tells us \emph{whether} the harness improved
at each iteration; it does not tell us \emph{which} edits were
associated with the improvement. To get at this, we examine every
adjacent-iteration transition (\emph{previous}$\to$\emph{next}) on a
fixed task and ask:
when a particular atomic shell operation appears in the harness output
for the first time (absent in the previous iteration's harness output,
present in the next), is reward more likely to go up than the
all-transition baseline? Table~\ref{tab:op_lift} reports, for each
frequently-newly-added operation, the number of newly-added
occurrences ($n_{\text{add}}$), the number of those followed by a
reward increase ($n_{\text{pos}}$), the resulting positive-transition
rate, the lift of that rate over the all-transition baseline (in
percentage points), and the same lift expressed as a relative change.
Operations such as \texttt{cat}, \texttt{sed}, \texttt{which}, and
\texttt{test} are strongly associated with reward improvement (lift
ranging from $+17$ to $+60$~pp), suggesting that they often appear in
transitions that repair inspection or condition-checking gaps. By
contrast, \texttt{curl}, \texttt{echo}, and \texttt{grep} are weakly or
negatively associated in this analysis, indicating that their appearance
is less predictive of reward-improving transitions. This kind of operation-level
decomposition is only possible because the dual-layer experience bank
stores per-case execution traces alongside the scalar score, and it
provides a direct signal for which directions the search policy
should be biased toward in subsequent iterations.

\begin{table}[t]
\centering
\caption{Operation-level lift analysis on adjacent-iteration transitions.
The unit of observation is a single \emph{previous}$\to$\emph{next}
iteration pair on a fixed task in which a given atomic operation was
\emph{newly added} (absent before, present after).
$n_{\text{add}}$: newly-added occurrences;
$n_{\text{pos}}$: occurrences followed by a reward increase;
\emph{Pos.\ rate}: $n_{\text{pos}}/n_{\text{add}}$;
\emph{Lift (pp)}: \emph{Pos.\ rate} minus the all-transition baseline rate
($\sim$$13.2\%$), in percentage points;
\emph{Rel.\ lift}: the same lift relative to the baseline.
\textcolor{liftpositive}{Green}: above-baseline lift;
\textcolor{liftnegative}{red}: below-baseline.}
\label{tab:op_lift}
\footnotesize
\setlength{\tabcolsep}{6pt}
\renewcommand{\arraystretch}{1.10}
\arrayrulecolor{tableline}
\begin{tabular}{@{}lrrrrr@{}}
\toprule
\rowcolor{tableheadbg}
\tablehead{Operation} & \tablehead{$n_{\text{add}}$} & \tablehead{$n_{\text{pos}}$} &
\tablehead{Pos.\ rate} & \tablehead{Lift (pp)} & \tablehead{Rel.\ lift} \\
\midrule
\rowcolor{tablestripebg}
\texttt{cat}        & 11 &  8 & 72.7\% & \textcolor{liftpositive}{$+59.5$} & \textcolor{liftpositive}{$+451.9\%$} \\
\texttt{sed}        & 11 &  4 & 36.4\% & \textcolor{liftpositive}{$+23.2$} & \textcolor{liftpositive}{$+175.9\%$} \\
\rowcolor{tablestripebg}
\texttt{which}      & 15 &  5 & 33.3\% & \textcolor{liftpositive}{$+20.2$} & \textcolor{liftpositive}{$+152.9\%$} \\
\texttt{test}       & 46 & 14 & 30.4\% & \textcolor{liftpositive}{$+17.3$} & \textcolor{liftpositive}{$+130.9\%$} \\
\rowcolor{tablestripebg}
\texttt{pip}        & 10 &  3 & 30.0\% & \textcolor{liftpositive}{$+16.8$} & \textcolor{liftpositive}{$+127.6\%$} \\
\texttt{python3}    & 19 &  5 & 26.3\% & \textcolor{liftpositive}{$+13.1$} & \textcolor{liftpositive}{$+99.7\%$}  \\
\rowcolor{tablestripebg}
\texttt{strings}    &  4 &  1 & 25.0\% & \textcolor{liftpositive}{$+11.8$} & \textcolor{liftpositive}{$+89.7\%$}  \\
\texttt{pdftotext}  & 15 &  3 & 20.0\% & \textcolor{liftpositive}{$+6.8$}  & \textcolor{liftpositive}{$+51.8\%$}  \\
\rowcolor{tablestripebg}
\texttt{head}       &  5 &  1 & 20.0\% & \textcolor{liftpositive}{$+6.8$}  & \textcolor{liftpositive}{$+51.8\%$}  \\
\midrule
\texttt{grep}       &  9 &  1 & 11.1\% & \textcolor{liftnegative}{$-2.1$}  & \textcolor{liftnegative}{$-15.7\%$}  \\
\rowcolor{tablestripebg}
\texttt{echo}       & 28 &  3 & 10.7\% & \textcolor{liftnegative}{$-2.5$}  & \textcolor{liftnegative}{$-18.7\%$}  \\
\texttt{curl}       & 19 &  1 &  5.3\% & \textcolor{liftnegative}{$-7.9$}  & \textcolor{liftnegative}{$-60.1\%$}  \\
\rowcolor{tablestripebg}
\texttt{file}       &  3 &  0 &  0.0\% & \textcolor{liftnegative}{$-13.2$} & \textcolor{liftnegative}{$-100.0\%$} \\
\texttt{jq}         &  3 &  0 &  0.0\% & \textcolor{liftnegative}{$-13.2$} & \textcolor{liftnegative}{$-100.0\%$} \\
\rowcolor{tablestripebg}
\texttt{rg}         &  3 &  0 &  0.0\% & \textcolor{liftnegative}{$-13.2$} & \textcolor{liftnegative}{$-100.0\%$} \\
\texttt{apt-get}    &  2 &  0 &  0.0\% & \textcolor{liftnegative}{$-13.2$} & \textcolor{liftnegative}{$-100.0\%$} \\
\rowcolor{tablestripebg}
\texttt{wc}         &  2 &  0 &  0.0\% & \textcolor{liftnegative}{$-13.2$} & \textcolor{liftnegative}{$-100.0\%$} \\
\texttt{pip3}       &  2 &  0 &  0.0\% & \textcolor{liftnegative}{$-13.2$} & \textcolor{liftnegative}{$-100.0\%$} \\
\bottomrule
\end{tabular}
\arrayrulecolor{black}
\vspace{-15pt}
\end{table}

\end{document}